\pdfoutput=1
\documentclass[11pt]{article}

\usepackage[final]{acl}

\usepackage{times}
\usepackage{latexsym}
\usepackage{amsmath}
\usepackage{amssymb}
\usepackage{booktabs}
\usepackage{multirow}
\usepackage{multicol}
\usepackage{makecell}
\usepackage{CJKutf8}
\usepackage{graphicx}
\usepackage{subfigure}
\usepackage{url}
\usepackage[T1]{fontenc}
\usepackage[utf8]{inputenc}

\usepackage{microtype}

\usepackage{inconsolata}

\usepackage{graphicx}

\title{Learn More, Forget Less: A Gradient-Aware Data Selection Approach for LLM}

\author{
 \textbf{Yibai Liu\textsuperscript{1}}\footnotemark[1],
 \textbf{Shihang Wang\textsuperscript{1}},
 \textbf{Zeming Liu\textsuperscript{2}}\footnotemark[2],
 \textbf{Zheming Song\textsuperscript{2}},
\\
 \textbf{Junzhe Wang\textsuperscript{2}},
 \textbf{Jingjing Liu\textsuperscript{2}},
 \textbf{Qingjie Liu\textsuperscript{2}},
 \textbf{Yunhong Wang\textsuperscript{2}}
\\
\textit{yl4616@columbia.edu, zmliu@buaa.edu.cn}\\
\small{\textsuperscript{1}Fu Foundation School of Engineering and Applied Science, Columbia University}\\
\small{\textsuperscript{2}School of Computer Science and Engineering, Beihang University}\\
}

\begin{document}
\maketitle
\begin{abstract}

Despite large language models (LLMs) have achieved impressive achievements across numerous tasks, supervised fine-tuning (SFT) remains essential for adapting these models to specialized domains. However, SFT for domain specialization can be resource-intensive and sometimes leads to a deterioration in performance over general capabilities due to catastrophic forgetting (CF). To address these issues, we propose a self-adaptive gradient-aware data selection approach (GrADS) for supervised fine-tuning of LLMs, which identifies effective subsets of training data by analyzing gradients obtained from a preliminary training phase. Specifically, we design self-guided criteria that leverage the magnitude and statistical distribution of gradients to prioritize examples that contribute the most to the model's learning process. This approach enables the acquisition of representative samples that enhance LLMs understanding of domain-specific tasks. Through extensive experimentation with various LLMs across diverse domains such as medicine, law, and finance, GrADS has demonstrated significant efficiency and cost-effectiveness. Remarkably, utilizing merely 5\% of the selected GrADS data, LLMs already surpass the performance of those fine-tuned on the entire dataset, and increasing to 50\% of the data results in significant improvements! With catastrophic forgetting substantially mitigated simultaneously. We will release our code for GrADS later.
\end{abstract}

\section{Introduction}

Although LLMs have achieved remarkable performances in multiple tasks such as open-domain question-answering \cite{achiam2023gpt, yang2024qwen2}, logical inference \cite{nam2024using}, and long-context understanding \cite{chen2023longlora}, supervised fine-tuning remains indispensable for domain-specific scenarios \cite{chen2023disc,yue2023disclawllm,xiong2023doctorglm,yang2023mentalllama}. However, 
% even with PEFT (Parameter-Efficient Fine-Tuning) methods \cite{lora, dettmers2024qlora}, 
incorporating domain-specific knowledge and concepts into the LLM parameters could be rather costly. For the sake of efficiency promotion, some studies have shown that not all fine-tuning data are useful \cite{zhou2024lima}, and removing some of the low-quality data instead can enhance model performance \cite{chen2023alpagasus, li2023quantity, cao2023instruction}.

\begin{figure}[t]
    \centering
   \includegraphics[width=\linewidth]{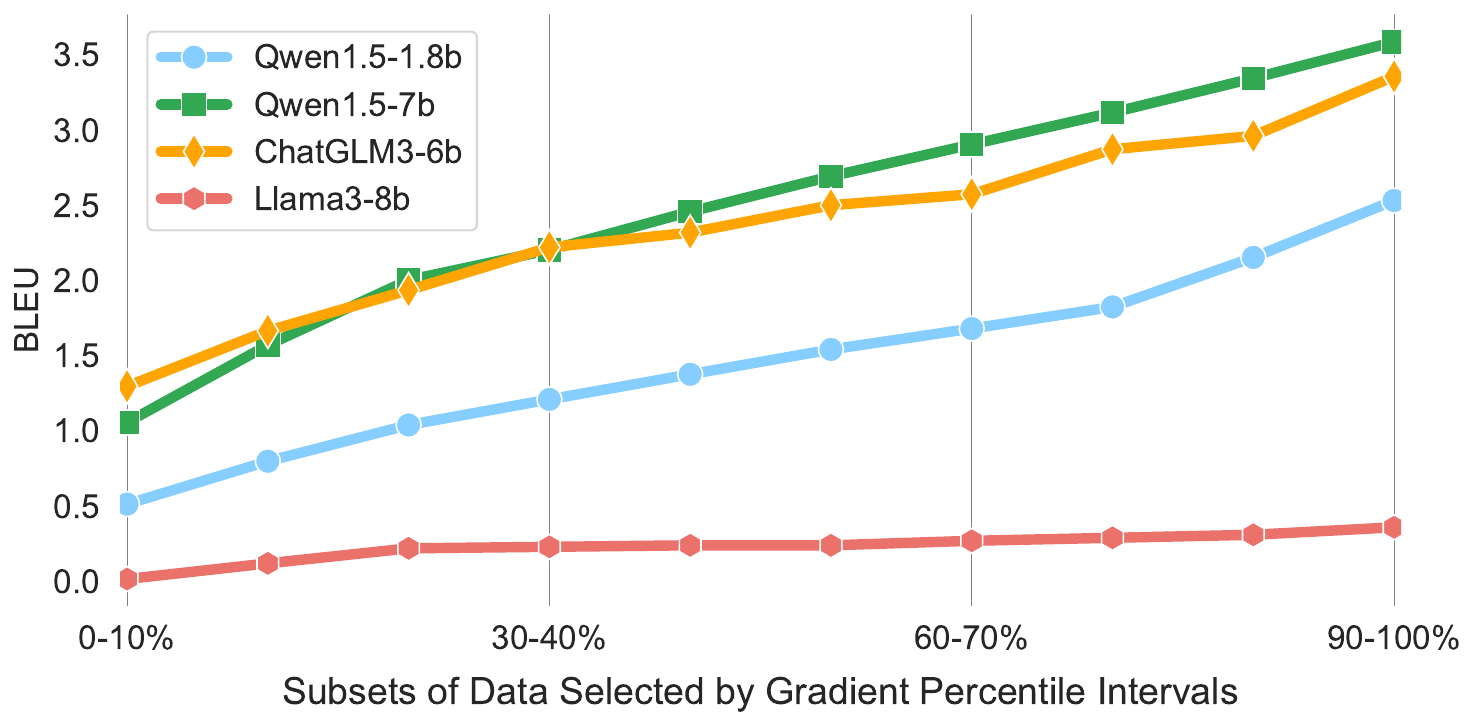}
    \caption{Pilot study: From left to right on the x-axis, we sort the CMedQA training data by gradients from largest to smallest, and select 10\% by rank at each time, conducting 10 subsets, and predict their responses with untuned LLMs.}
    \label{fig:pilot study}
\end{figure}

Besides, after domain-oriented fine-tuning, LLMs typically experience a decline in general capabilities, a phenomenon named Catastrophic Forgetting (CF) \cite{kaushik2021understanding, cossu2022continual}. To address this issue, some practices use a mixture of domain-specific and general data \cite{luo2024empirical}, and others propose additional regularization or adaptation techniques \cite{ ke2024continual,diao2023mixture}. However, these approaches either increase the computational cost or compromise domain expertise to preserve more general capabilities \cite{lin2023speciality}. 

To promote training efficiency and mitigate CF, we focus on leveraging LLMs to select high-quality subsets of data for training. Analogous to a well-educated student who can discern the most suitable college courses through trial classes, we posit that a sufficiently pre-trained LLM is capable of identifying data that is more beneficial for its learning during the fine-tuning phase. Inspired by past works estimating the influence of training instances with gradient information \cite{pruthi2020estimating, han2023understanding, xia2024less}, we design a gradient-aware approach to select such data.

Therefore, we conduct a pilot study that illustrates the performance of vanilla LLMs in predicting outputs for slices of training data, each selected from different gradient intervals (Figure \ref{fig:pilot study}). The results show that the LLMs have higher accuracy when predicting data characterized by smaller gradients (right side) as opposed to larger gradients (left side), which confirms the potential of gradients in training data selection.

To effectively identify crucial data from candidate training sets, we propose an adaptive \textbf{Gradient-Aware Data Selection} method, namely \textbf{GrADS}. First, the entire candidate data would be trained for one epoch with the LLMs to extract gradients for each training instance. Then, a self-adaptive criterion based on the gradient distribution is used to select a subset from the full data of expected volume. This method eschews reliance on expensive, more advanced LLMs like GPT-4 for inference \cite{chen2023alpagasus, du2023mods, liu2023makes} and the need for manual intervention in creating high-quality seed data \cite{pan2024g, ge2024clustering}, thereby offering a cost-effective and pragmatic solution.

To carefully examine the effectiveness of GrADS, we conducted comprehensive experiments on various LLMs including Qwen \cite{qwen}, ChatGLM \cite{zeng2023glm-130b}, and Llama \cite{llama3modelcard} scaled from 1.8B, around 7B, to 14B, within typical knowledge-intensive and high-demand application domains like medicine \cite{huatuo}, law \cite{chatlaw}, and finance \cite{xuanyuan}. GrADS exhibits superior advantages in terms of efficiency, cost-effectiveness, and performance. In summary, our contributions are three folds:

\begin{itemize}
    \item We introduce a novel self-adaptive Gradient-Aware Data Selection method (GrADS), which operates independently of manual intervention. 
    \item Extensive experiments across different LLMs, model scales, and domains validate the efficacy of GrADS in facilitating target task performance.
    \item GrADS substantially mitigates the catastrophic forgetting problem, achieving an outstanding balance between domain specialization and general capabilities.

\end{itemize}

\section{Related Work}

\subsection{Data Selection}
The recent research by Zhou et al.\cite{zhou2024lima} indicates that most of the knowledge in LLMs is acquired during the pre-training phase, and a limited amount of instruction data is often sufficient to activate the models' capacity to follow instructions. Similarly, through interactions with SoTA LLMs such as GPT-4, Chen et al. \cite{chen2023alpagasus} introduced ALPAGASUS, while Li et al. \cite{li2023quantity} proposed the Instruction-Following Difficulty (IFD) metric to select samples with desired characteristics to enhance LLM instruction tuning. Liu et al. \cite{liu2023makes} and Du et al. \cite{du2023mods} further delineated a series of criteria including quality, complexity, diversity, coverage, and necessity to select data. Additionally, some researchers constructed expert-aligned datasets \cite{ge2024clustering} or curated high-quality seed data \cite{pan2024g}. By facilitating interaction between the LLM and these datasets alongside the original data, they aim to obtain feedback on the quality of the data and improve the models' performance.

However, the majority of the previous works necessitate human intervention or the involvement of SoTA LLMs such as GPT-4 for data filtering, which require substantial API quota budgets or significant human labor investments. In contrast, our data selection method addresses efficiency and cost-effectiveness, which is easy to implement and substantially reduces labor and API expenditures.

\subsection{Catastrophic Forgetting}
Domain-specific fine-tuned LLMs \cite{ouyang2022rlhf, hyung2024scaling} have demonstrated substantial potential for knowledge-based question-answering (QA), auxiliary consulting, and personalized solution recommendation in various fields, such as medicine\cite{huatuo}, finance\cite{xuanyuan} and law \cite{chatlaw}. However, as expertise within the domain advances, CF emerges along with a sharp decline in the general capabilities that the LLM had previously mastered during pre-training \cite{kaushik2021understanding, cossu2022continual, luo2024empirical}. 

To address this issue, from a data-driven perspective, some researchers trained both domain data and general data to reduce the forgetting of general knowledge \cite{chen2020recall}, while others proposed self-distillation \cite{yang2024self}, which guides the generation of task data through the model itself to minimize the disparity between the information distribution of the generated data and that of the initial model. On the other hand, from the model's perspective, some established end-to-end alignment of modules through shared attention mechanisms \cite{zhao2024sapt}, while others modified the adapter architecture by reducing the interference caused by fine-tuning tasks in different orthogonal low-rank subspaces \cite{wang2023olora} or by self-regulating the adapter's attention to different parts of the context \cite{liu2024more}.

% Nonetheless, data-driven methods often involve extensive experimentation to determine appropriate instances, mixing ratios, and training orders, thereby incurring greater computational resource costs. Model-driven methods, on the other hand, frequently compromise domain specialization to preserve more general capabilities. In comparison, our proposed method is not only computationally efficient but also capable of mitigating CF without compromising domain performance.

\section{Backgrounds}

The \textbf{Embedding} layer and the \textbf{language model head (LM Head)} layer of LLMs play critical roles in capturing the semantics of input tokens and generating meaningful predictions, respectively. The Embedding layer maps each discrete token into a high-dimensional vector space, where the vectors capture the semantic and syntactic properties of the words they represent. On the other hand, the LM Head layer converts the final hidden states produced by the model into a probability distribution over the vocabulary and directly influences the model's accuracy in predicting the next token.

During back-propagation, the gradients computed for the Embedding layer indicate how the word vectors need to be updated to optimize the discriminative and context-aware token representations, which enhance the model's overall performance. Therefore, instances with larger gradients for the Embedding layer could imply the existence of unfamiliar information the model attempts to learn, while those with smaller gradients are rather stable and already well-presented.

In contrast, the gradients computed for the LM Head provide insights into how the model should adjust its parameters to minimize the prediction error in the decoding process, thereby improving its predictive capabilities. High-magnitude gradients show uncertainty and lack of confidence in the model's prediction, which reveals potential high complexity and perplexity of the data, whereas low-gradient tokens are well-understood and straight forward to the model.

With the insight that gradients help discover characteristics of each training instance, we raise a deduction that in a given training dataset $D$, the actual "effective" data points $D'$ should depend on $feature$ $importance$ ($F$), $information$ $values$ ($I$), and $complexity$ ($C$):

\begin{equation}
D' \propto f(F, I, C)
\end{equation}

\section{GrADS: Gradient-Aware Data Selection}

\begin{figure*}[!ht]
    \centering
    \includegraphics[width=1.0\linewidth]{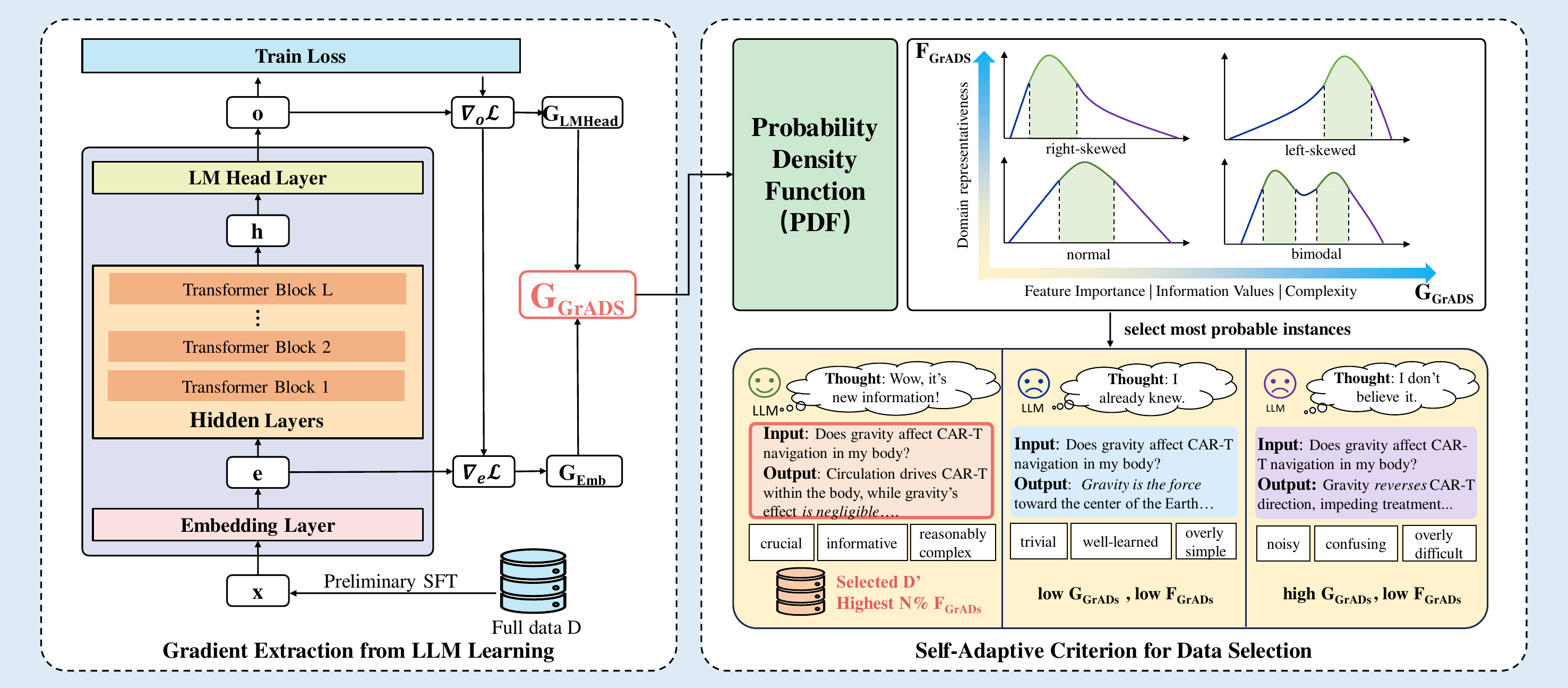}
    \caption{The illustration of the GrADS method.}
    \label{fig:illustration of GrADS}
\end{figure*}

In this section, we introduce GrADS, which can adaptively select beneficial subsets of the data through gradient distribution, integrating both the Embedding layer gradients and LM Head gradients. The method consists of two major steps: gradient extraction from LLM learning, and data selection with a self-adaptive criterion. Specifically, in the first step, we obtain the integrated gradients of each training instance by conducting a single-epoch SFT. Subsequently, we select desired subsets of data based on an adaptive criterion derived from the gradient distribution. Our model architecture is illustrated in Figure \ref{fig:illustration of GrADS}.

\subsection{Gradient Extraction from LLM Learning}

Given the entire training data $\mathbf{D}$, we denote the input tokens of each data point $\mathbf{x} = \{x_1, x_2, \ldots, x_T\} \in \mathbb{R}^{T}$, where $T$ is the length of the input sequence. In the Transformer embedding layer, tokens are mapped to the corresponding embedding vectors $\mathbf{e}$:
\begin{equation}
\mathbf{e} = Embed(\mathbf{x})
\end{equation}
where $\mathbf{e} = \{\mathbf{e}_1, \mathbf{e}_2, \ldots, \mathbf{e}_T\} \in \mathbb{R}^{T*d}$ is the combined vector for input tokens, $d$ denoting the dimension.

Then the embedded vectors $\mathbf{e}$ are passed through multiple Transformer layers, and produce the final hidden states $\mathbf{h} \in \mathbb{R}^{T*d}$ for all training instances.
\begin{equation}
\mathbf{h} = Transformer(\mathbf{e})
\end{equation}

The LM Head layer takes the final hidden states $\mathbf{h}$ and generates the probability distribution over the vocabulary for predicting the next token:
\begin{equation}
\mathbf{o} = softmax(LM Head(\mathbf{h}))
\end{equation}
where $\mathbf{o} \in \mathbb{R}^{T*V}$ is the probability distribution over the vocabulary for the next token, and $V$ is the size of the vocabulary. 

Given a standard cross entropy loss function $\mathcal{L}(\cdot)$ that measures the difference between the model's predictions and the ground truth, the gradients of the loss for the embeddings can be obtained in the forward pass by:

\begin{equation}\label{eq:grad-embed}
\mathbf{g}_{\text{Emb}} = \nabla_{\mathbf{e}} \mathcal{L} = \left(\frac{\partial \mathbf{h}}{\partial \mathbf{e}}\right)^{\top} \nabla_{\mathbf{h}} \mathcal{L}
\end{equation}
where $\nabla_{\mathbf{h}} \mathcal{L}$ is the gradient of the loss for the hidden state $\mathbf{h}$ and $\frac{\partial \mathbf{h}}{\partial \mathbf{e}}$ represents the Jacobian matrix \cite{wilamowski2008computing} of the hidden state for the embedding. The gradients $\nabla_{\mathbf{h}} \mathcal{L}$ can be obtained through backpropagation through the model.

Similarly, we can compute the gradients for the LM Head layer during the back-propagation step:

\begin{equation}\label{eq:grad-lmhead}
\mathbf{g}_{\text{LM}} = \nabla_{\mathbf{o}} \mathcal{L} = \frac{\partial \mathcal{L}}{\partial \mathbf{o}}
\end{equation}

After extracting the gradients for the input tokens in the Embedding and LM Head layers, we exclude special tokens like $[CLS]$, $[SEP]$, $[PAD]$, $[UNK]$, etc. Since the gradients of the Embedding layer reflect the LLMs' understanding of the input sequence whereas the gradients of the LM Head layer reflect the LLMs' certainty of the output tokens, we take all input tokens for the Embedding layer and only the output token for the LM Head layer into account. Meanwhile, to rule out the impact of input sequence length, we average the token-wise gradients for each training instance. Thus, the combined instance-level gradients for two layers are denoted as $\mathbf{G}^{i}_{\text{Emb}}$ and $\mathbf{G}^{i}_{\text{LM}}$, where $\mathbf{i} = \{\mathbf{1}, \mathbf{2}, \ldots, \mathbf{K}\}$ and $\mathbf{K}$ being the size of the data $\mathbf{D}$. 

Thereafter, we integrate these gradients by adding $\mathbf{G}_{\text{Emb}}$ and $\mathbf{G}_{\text{LM}}$ linearly to derive a GrADS gradient vector $\mathbf{G}_{\text{GrADS}}$, which ultimately serves as the metric for selecting training instances

% Equation for the combined gradient
\begin{equation}
G_{\text{GrADS}} =G_{\text{Emb}} + G_{\text{LM}}
\end{equation}

\subsection{Self-Adaptive Criterion for Data Selection}

To select the subset of training instances that best represents domain knowledge, we introduce the probability density function (PDF) to depict the distribution feature of $\mathbf{G}_{\text{GrADS}}$. The PDF uses a non-parametric method, such as kernel density estimation (KDE) to represent the density of $\mathbf{G}_{\text{GrADS}}$ at different values. A higher density signifies that there are more instances whose $\mathbf{G}_{\text{GrADS}}$ fall close, indicating instances more likely to share common domain characteristics. Just as one can quickly gain an understanding of a domain by reading its classical papers, prioritizing the fine-tuning process on these typical training instances can also enable LLMs to learn domain knowledge more efficiently and effectively.

Therefore, we compute the PDF function of $\mathbf{G}_{\text{GrADS}}$ gradients:

\begin{equation}
F_{\text{GrADS}} = PDF(G_{\text{GrADS}})
\end{equation}
where $\mathbf{F_{\text{GrADS}}} \in \mathbb{R}^{\mathbf{K}}$ implies the domain representativeness of the instance. Finally, an effectively refined subset of the full data $\mathbf{D}$ can be obtained by calculating the Top N\% of instances with the highest PDF values:

% Equation for the combined gradient
\begin{equation}
D^{'} = \text{quantile}\left(F_{\text{GrADS}}, N / 100\right)
\end{equation}

GrADS operates in a self-adaptive manner, as it selects the most probable training instances regardless of the gradient distribution, whether it is left-skewed, right-skewed, normal, bimodal, or otherwise. The selected subset $\mathbf{D^{'}}$ always has the highest $\mathbf{F_{\text{GrADS}}}$, thereby best capturing the critical characteristics of the domain. Also, $\mathbf{D^{'}}$ carries crucial, informative, and reasonably challenging instances that guide the model to learn and acquire domain expertise. Nonetheless, as Figure \ref{fig:illustration of GrADS} illustrates, training instances with low $\mathbf{G_{\text{GrADS}}}$ and low $\mathbf{F_{\text{GrADS}}}$ are typically less representative, often characterized as trivial, well-learned, or simple, and thus fail to "surprise" the model with already-known knowledge. In contrast, instances with high $\mathbf{G_{\text{GrADS}}}$ and low $\mathbf{F_{\text{GrADS}}}$ are often noisy, confusing, or overly difficult, and they might introduce misleading information that contradicts the model's established common sense. GrADS excludes these suboptimal instances by automatically adapting to the distribution of domains.

\section{Experiments}

\begin{table*}[!ht]
    \centering
    \footnotesize
    \begin{tabular}{cl|m{0.8cm}<{\centering}m{0.8cm}<{\centering}m{1.1cm}<{\centering}|m{0.8cm}<{\centering}m{0.8cm}<{\centering}m{1.1cm}<{\centering}|m{0.8cm}<{\centering}m{0.8cm}<{\centering}m{1.1cm}<{\centering}}
    % \begin{tabular}{cl|ccc|ccc|ccc}
    \toprule
         \multirow{2}{*}{\textbf{Base Model}}& \multirow{2}{*}{\textbf{Method}} & \multicolumn{3}{c|}{\textbf{CMedQA}} & \multicolumn{3}{c|}{\textbf{LawQA}} & \multicolumn{3}{c}{\textbf{FinQA}} \\
         &  & BLEU & ROUGE & METEOR & BLEU & ROUGE &	METEOR & BLEU & ROUGE & METEOR \\
         \midrule
\multirow{7}{*}{\shortstack{Qwen1.5-7B}}&base&2.627& 12.180 & 10.860 & 9.066 & 20.050 & 21.392 & 3.188 & 11.194 & 14.669 \\
&all & 3.813 & 17.327 & 12.276 & 16.090 & 27.603 & 27.472 & 10.120 & 24.067 & 18.757\\
\cmidrule(lr){2-11} 
&rdn & 3.548 & 16.776 & 11.954 & 15.856 & 27.288 & 26.810 & 9.686 & 22.621 & 17.276\\
&bm25 & 4.133 & 18.152 & 13.260 & 16.667 & 27.538 & 28.264 & 10.419 & 23.837 & 20.645\\
&dsir & 3.650 & 17.636 & 12.314 & 15.987 & 27.362 & 27.644 & 9.876 & 23.463 & 19.142\\
&rds & 3.826 & 17.980 & 12.744 & 16.203 & 27.862 & 28.017 & 10.133 & 24.135 & 20.075\\
&ppl & 4.871 & 18.285 & 14.689 & 18.013 & 27.776 & 30.660 & 11.419 & 23.304 & 23.325 \\
&less & \underline{5.126} & \underline{18.214} & \underline{14.896} & \underline{19.473} & \textbf{29.314} & \underline{33.727} & \underline{12.884} & \textbf{25.135} & \underline{23.976} \\
&\textbf{grads} & \textbf{5.372} & \textbf{18.496} & \textbf{15.396} & \textbf{20.270} & \underline{28.026} & \textbf{35.985} & \textbf{13.364} & \underline{24.822} & \textbf{24.872} \\
\midrule
\multirow{7}{*}{\shortstack{ChatGLM3-6B}}&base&2.568 & 11.274 & 15.634 & 7.966 & 19.733 & 19.011 & 3.174 & 11.437 & 14.926 \\
&all & 4.297 & 17.432 & 16.722 & 16.673 & 28.016 & 28.519 & 11.454 & 22.918 & 24.898\\
\cmidrule(lr){2-11} 
&rdn & 4.512 & 16.674 & 16.482 & 16.453 & 27.576 & 27.864 & 11.216 & 22.450 & 24.233\\
&bm25 & 4.824 & 17.015 & 17.163 & 16.929 & 27.798 & 28.316 & 11.636 & 23.412 & 25.170\\
&dsir & 4.330 & 16.488 & 15.856 & 16.215 & 26.943 & 27.534 & 11.328 & 22.390 & 24.421\\
&rds & 4.607 & 17.216 & 16.754 & 17.036 & 27.689 & 28.525 & 11.596 & 23.538 & 24.427\\
&ppl & 5.031 & 17.503 & 17.637 & 18.865 & \textbf{28.411} & 33.068 & 11.957 & \underline{24.214} & \underline{26.682} \\
&less & 5.283 & \textbf{18.425} & \textbf{18.529} & \underline{19.002} & \underline{28.214} & \underline{33.337} & \underline{12.216} & 23.790 & 26.394 \\
&\textbf{grads} & \textbf{5.488} & \underline{17.813} & \underline{18.375} & \textbf{20.288} & 28.067 & \textbf{34.932} & \textbf{13.165} & \textbf{24.281} & \textbf{28.567} \\
\midrule
\multirow{7}{*}{\shortstack{Llama3-8B}}&base& 0.026 & 0.249 & 0.291 & 0.259 & 1.905 & 2.164 & 0.178 & 1.293 & 1.225 \\
&all & 3.332 & 16.415 & 11.061 & 15.272 & 24.301 & 27.033 & 9.116 & 21.190 & 16.913\\
\cmidrule(lr){2-11} 
&rdn & 3.265 & 15.884 & 10.798 & 15.552 & 24.688 & 26.476 & 9.337 & 22.654 & 16.870 \\
&bm25 & 3.474 & 16.763 & 12.018 & 15.859 & 24.803 & 28.165 & 10.225 & 22.387 & 18.244 \\
&dsir & 3.206 & 15.817 & 11.001 & 14.643 & 24.112 & 25.386 & 9.640 & 22.818 & 17.266 \\
&rds & 3.399 & 16.352 & 12.679 & 15.704 & 24.638 & 27.766 & 10.413 & 22.694 & 18.375 \\
&ppl & 4.183 & \textbf{17.809} & 13.632 & 16.390 & \underline{25.122} & 30.378 & \underline{11.863} & \underline{22.817} & 22.469 \\
&less & \underline{4.213} & 17.130 & \underline{13.845} & \underline{16.737} & 25.015 & \underline{31.408} & 11.480 & 22.526 & \underline{23.425} \\
&\textbf{grads} & \textbf{4.472} & \underline{17.365} & \textbf{14.089} & \textbf{18.751} & \textbf{26.613} & \textbf{34.620} & \textbf{12.288} & \textbf{23.678} & \textbf{23.437} \\
\bottomrule
    \end{tabular}
    \caption{Main Results. \textit{base} denotes no further training implemented, \textit{all} denotes full dataset, and otherwise we select 50\% of the data for training.}
    \label{tab:main results}
\end{table*}

In this section, we present the experiment results to verify the effectiveness of GrADS. Apart from the main results, we also try to validate the generalizability of GrADS by addressing the following research questions (RQs):(1) Generalizability: Can the GrADS approach be scaled up to larger LLMs and applied across different models? (2) Robustness: Do GrADS consistently perform well with smaller subsets selected? 
 % (1) Transferability: Do data selected by GrADS using smaller LLMs also enhance the performance of larger-scale models? 
%(4) Can GrADS outperform subsets selected from different gradient spans?

\begin{table}[!ht]
    \centering
    \scriptsize
    % \footnotesize
    % \begin{tabular}{cl|c|c|c}
    \begin{tabular}{lc|m{0.9cm}<{\centering}|m{0.9cm}<{\centering}|m{0.9cm}<{\centering}}
    \toprule
         \textbf{Base Model}& \textbf{Method} & \textbf{CMedQA} & \textbf{LawQA} & \textbf{FinQA} \\
         \midrule
\multirow{2}{*}{\shortstack{Qwen1.5-7B}}& all & 2.712 & 3.318 & 2.679 \\
&\textbf{grads} & \textbf{3.159} & \textbf{4.202} & \textbf{3.295} \\
\midrule
\multirow{2}{*}{\shortstack{ChatGLM3-6B}}& all & 2.587 & 3.254 & 2.826 \\
&\textbf{grads} & \textbf{3.215} & \textbf{4.034} & \textbf{3.336} \\
\midrule
\multirow{2}{*}{\shortstack{Llama3-8B}}& all & 2.553 & 3.110 & 2.547 \\
&\textbf{grads} & \textbf{2.887} & \textbf{3.823} & \textbf{2.914} \\
\bottomrule
    \end{tabular}
    \caption{Results by GPT-4o's evaluation, scores range from 1-5. }
    \label{tab:gpt4o results}
\end{table}

\subsection{Datasets}
Our study incorporates three domains-specific datasets from three typical domains: CMedQA \cite{zhang2018multi} for medicine, LawQA \cite{huang2023lawyer} for law, and FinQA \footnote{https://aistudio.baidu.com/datasetdetail/34744} for finance. The CMedQA dataset is provided by qualified experts, the LawQA dataset is generated by advanced LLMs, and the FinQA dataset is sourced from the open web and undergone post-cleaning. These datasets encompass the primary methodologies for fine-tuning data collection currently used, making experimental conclusions derived from those datasets representative, and can be reasonably expected to generalize to a wider range of data.

Specifically, CMedQA includes 20k instances for training and 0.5k instances for testing. For LawQA, we use the law article-based QA pairs from the Lawyer-LLama project \cite{Lawyer-LLama} and split 1.6k and 0.4k data for training and testing, respectively. Since FinQA's QA pairs are sourced from webpages, we retained only those designated as "best answers" in the original dataset. Additionally, we removed all duplicate questions and answers, resulting in a training set of 40k and a testing set of 2k.

\subsection{Evaluation Metrics}
We follow Pan et al. \cite{pan2024g} to include BLEU \cite{papineni2002bleu}, along with ROUGE-L \cite{lin2004rouge} and METEOR \cite{banerjee2005meteor} to evaluate the response quality. In addition, we employed GPT-4o to score on a 1-5 scale of the response quality. We also provide the consistency test between GPT-4o evaluation and human judgement in Appendix A.

Furthermore, we delve into the CF problem in general capabilities following supervised fine-tuning on domain-specific instances. To this end, we follow the work of Liu et al. \cite{liu2024more} and collect C-Eval \cite{huang2023ceval} for common sense understanding, GSM8K \cite{yu2023metamath} for mathematics, ALPACA \cite{peng2023instruction} for instruction following and SafetyPrompts \cite{sun2023safety} for instruction attack and typical safety scenarios awareness.

For C-Eval, we write a rule-based method to extract the options predicted by LLMs, and report the accuracy and whether the LLMs follow the instruction of "Single-choice questions". For GSK8k, we apply its publicly released Chinese version which is translated by GPT3.5-Turbo \footnote{https://huggingface.co/datasets/meta-math/GSM8K\_zh}. We follow the previous work \footnote{https://github.com/QwenLM/Qwen} to extract the numerical results predicted by LLMs and report the accuracy, BLEU, and ROUGE-L. For ALPACA, we report BLEU and ROUGE-L. For SafetyPrompts (Typical Safety and Instruction Attack subdata), we write a few-shot prompt to instruct GPT-4o to conduct a 2 choice task on whether the LLMs' responses are safe or not. The responses are considered as correct if GPT-4o labels them as "safe". 

% Notably, due to computational limits, we randomly sampled 1k, 1k2, and 1k4 samples from ALPACA, instruction attack, and typical safety datasets respectively for testing. 

% We illustrate our main results in SubSection \ref{subsec: main results} and provide more indepth experiments and discussions in SubSection \ref{subsec: indepth}.

\subsection{Foundation Models}
To validate GrADS' efficiency across different model scales and model architectures, we selected Qwen1.5-7B-Chat \cite{qwen}, ChatGLM3-6B-Chat \cite{zeng2023glm-130b} and Llama3-8B-Instruct \cite{llama3modelcard} as our base LLMs. We also selected Qwen1.5-1.8B-Chat and Qwen1.5-14B-Chat for the RQ1 investigation. Besides, while GrADS permits any proportion of data selection from the original training sets, we uniformly select 50\% in the main experiments for simplicity, the exploration of varying proportions will be conducted in RQ2.

To have a thorough understanding of GrADS performance regarding different training methods, we implement full-parameter fine-tuning in our main results and investigate LoRA training in Appendix F. 

\subsection{Baselines}
Despite the existence of numerous data selection methods, we automatically excluded those requiring manual intervention \cite{pan2024g, ge2024clustering} or extensive use of advanced LLMs (like GPT-4) \cite{chen2023alpagasus, liu2023makes}. Consequently, we mainly follow the settings of Less \cite{xia2024less} and select \textbf{Random Selection}, \textbf{BM25} \cite{robertson2009probabilistic}, \textbf{DSIR} \cite{xie2023data}, \textbf{RDS} \cite{zhang2018unreasonable, hanawa2020evaluation}, \textbf{LESS} \cite{xia2024less} as baselines. The implementation of RDS also follows the setting in \citet{xia2024less}. Apart from the above methods, to validate the effectiveness of gradient in GrADS, we also replace gradient with perplexity score for each training instance, denoted as \textbf{PPL}. We have some further illustration regarding those baselines in Appendix G.

\begin{table*}[!ht]
    \centering
    \footnotesize
    \begin{tabular}{cl|cc|ccc|cc|c|c}
    \toprule
         \multirow{2}{*}{\textbf{Domain}}& \multirow{2}{*}{\textbf{Method}} & \multicolumn{2}{c|}{\textbf{C-Eval}} & \multicolumn{3}{c|}{\textbf{GSM8k}} & \multicolumn{2}{c|}{\textbf{ALPACA}} & \multicolumn{1}{c|}{\textbf{Safety}} & \multicolumn{1}{c}{\textbf{Attack}} \\
         &  & Acc. & Instruct & Acc. &	BLEU & ROUGE & BLEU & ROUGE & Acc. & Acc. \\
         \midrule
\multirow{6}{*}{\shortstack{CMedQA}}&base&65.189& 87.427 & 55.497 & 14.967 & 29.207 & 15.097 & 27.529 & 43.807 & 51.365 \\
\midrule
&all & 11.285 & 22.674 & 1.895 & 2.286 & 12.809 & 1.815 & 12.252 & 13.594 & 22.007 \\
%\cmidrule(lr){2-11} 
&rdn & 14.628 & \textbf{35.107} & 2.880 & 3.006 & 13.410 & 2.443 & 13.305 & 14.446 & 28.503 \\
%&top50 & 6.198 & 7.413 & 1.744 & 1.69 & 11.105 & 1.548 & 11.817 & 16.227 & 25.214 \\
%&tail50 & \textbf{23.764} & \underline{34.462} & \textbf{5.231} & \textbf{5.668} & \underline{18.060} & \textbf{3.788} & \underline{15.511} & \textbf{26.122} & \textbf{34.424} \\
&grads & \textbf{21.345} & 33.293 & \textbf{4.700} & \textbf{5.572} & \textbf{18.082} & \textbf{3.340} & \textbf{15.626} & \textbf{24.560} & \textbf{33.656} \\
\midrule
\multirow{3}{*}{\shortstack{LawQA}}
&all & 24.201 & 8.067 & 15.466 & 9.985 & 20.917 & 8.987 & 20.055 & 23.571 & 39.961\\
%\cmidrule(lr){2-11} 
&rdn & 30.305 & 11.846 & 26.384 & 10.772 & 22.049 & 9.486 & 19.889 & 29.378 & 44.938\\
%&top50 & 20.203 & 6.178 & 20.925 & 10.563 & 21.608 & 8.389 & 18.806 & 16.932 & 30.039 \\
%&tail50 & 28.273 & \underline{12.192} & \textbf{26.990} & \textbf{11.165} & \underline{22.306} & \textbf{9.926} & \textbf{20.264} & \textbf{34.006} & \underline{47.089} \\
&grads & \textbf{31.206} & \textbf{12.762} & \textbf{26.547} & \textbf{10.913} & \textbf{22.368} & \textbf{9.506} & \textbf{20.145} & \textbf{32.596} & \textbf{48.452} \\
\midrule
\multirow{3}{*}{\shortstack{FinQA}}&all & 10.756 & 21.802 & 0.758 & 0.855 & 8.266 & 0.895 & 8.864 & 6.480 & 10.921\\
%\cmidrule(lr){2-11} 
&rdn & 15.77 & 28.488 & 0.758 & 0.795 & 7.665 & 1.073 & 9.563 & 9.688 & 15.141 \\
%&top50 & 8.503 & 14.026 & 1.060 & 0.564 & 7.610 & 0.850 & 9.001 & 7.764 & 12.497 \\
%&tail50 & \textbf{26.701} & \textbf{50.232} & \textbf{1.516} & \underline{2.198} & \underline{10.685} & \textbf{2.261} & \textbf{11.594} & \underline{17.741} & \textbf{26.694} \\
&grads & \textbf{25.250} & \textbf{47.359} & \textbf{1.373} & \textbf{2.416} & \textbf{10.838} & \textbf{1.873} & \textbf{10.974} & \textbf{19.139} & \textbf{25.889} \\
\bottomrule
    \end{tabular}
    \caption{Catastrophic forgetting results of Qwen1.5-7B. We select 50\% of data for \textit{rdn} and \textit{grads}.}
    \label{tab: main cf results}
\end{table*}

\begin{table}[!ht]
    \centering
    \scriptsize
    % \begin{tabular}{l|ccc|ccc}
    \begin{tabular}{l|m{0.60cm}<{\centering}m{0.60cm}<{\centering}m{0.80cm}<{\centering}|m{0.60cm}<{\centering}m{0.60cm}<{\centering}m{0.80cm}<{\centering}}
    \toprule
         \multirow{2}{*}{\textbf{Method}} & \multicolumn{3}{c|}{\textbf{Qwen1.5-1.8B Gradients}} & \multicolumn{3}{c}{\textbf{Qwen1.5-14B Gradients}} \\
         & BLEU & ROUGE & METEOR & BLEU & ROUGE & METEOR \\
         \midrule
        base &3.308 & 11.443 & 14.954 & 3.308 & 11.443 & 14.954 \\
        all & 12.518 & \underline{25.451} & 20.183 & 12.518 & \underline{25.451} & 20.183 \\
        \midrule
        rdn &10.674 & 23.758 & 18.337 & 10.674 & 23.758 & 18.337 \\
        ppl & 11.913 & 24.272 & 22.136 & 12.370 & 24.854 & 21.348 \\
        less & \underline{13.549} & 24.877 & \underline{24.863} & \underline{13.838} & 24.895 & \underline{24.572}\\
        grads & \textbf{14.169} & \textbf{25.844} & \textbf{25.739} & \textbf{14.371} & \textbf{25.925} & \textbf{26.673} \\
        \bottomrule
    \end{tabular}
    \caption{The left side selects data via Qwen1.5-1.8B gradients and fine-tuned on Qwen1.5-14B. The right side is selected via Qwen1.5-14B and fine-tuned on Qwen1.5-14B. The \textit{base}, \textit{all}, and \textit{rdn} are all based on Qwen1.5-14B, so they share the same results. We select 50\% of the data for \textit{rdn}, \textit{ppl}, \textit{less}, and \textit{grads}. }
    \label{tab: 14b results}
\end{table}

\subsection{Main Results}\label{subsec: main results}

\subsubsection{Domain Performance}

%We provide comparisons between the domain performances for base LLMs fine-tuned on different baseline datasets in Table \ref{tab:main results} and Table \ref{tab:gpt4o results}, evaluated by traditional automatic metrics and the response quality scores assessed by GPT-4o, respectively. 
Results in Table \ref{tab:main results} and Table \ref{tab:gpt4o results} show that (1) \textbf{GrADS obtained the best or second-best performance over almost all domains in the experiments.} Notably, with only 50\% of the data, GrADS has achieved remarkable improvements on BLEU and METEOR metrics, registering an average gain of 28.08\% and 25.57\% respectively, compared to LLMs fine-tuned on the entire dataset. Considering that the question-answering tasks require domain expertise, the higher BLEU, and METEOR indicate that the LLMs advance in both accuracy and richness of professional expression. Apart from that, the improvement on ROUGE-L also indicates that the LLMs have considerable enhancements in terms of long-sequence content coherence and comprehensive information coverage.

%Meanwhile, (2) we found that GrADS has the highest average improvement on BLEU values for various LLMs in the medical field, reaching 30.4\% and 4.7\%, respectively. This phenomenon meets our common expectations for the medical field, which is to have denser professional terminology, higher long sequence consistency, and more comprehensive content coverage. At the same time, GrADS has achieved the highest average improvement in METEOR values on the LLMS in the legal domain, reaching 27\%. This also satisfies our demand for an excellent legal consulting LLM that considers semantic similarity while balancing accuracy and recall, with special emphasis on the lexical and semantic relevance between generated text and reference text.

Besides, we found that (2) \textbf{GrADS is not sensitive to model initialization and model architecture, demonstrating very strong robustness.} Although Llama3 was mainly pre-trained on English datasets and perform poorly when it comes to Chinese set (as the \textit{base} experiment of Llama3 indicates), GrADS also substantially improve its performance as what it did for those well-pretrained Chinese background LLMs (Qwen and ChatGLM). 

Meanwhile, (3) \textbf{GrADS has attention on the domain characteristics when selecting data, namely self-adaptive}. When we sort all training instances by gradient magnitude in ascending order, the average percentiles of selected data for CMedQA, LawQA, and FinQA are 35.8\%, 27.4\%, and 28.9\%, respectively. In more specialized domains such as medicine where all base LLMs perform poorly, GrADS inclines to select instances with larger gradients (the harder ones).
% , indicating that the base LLMs require more knowledge-intensive learning to be a medical expert. 

\subsubsection{Catastrophic Forgetting}
To keep the paper reasonably concise, we only present the results of Qwen1.5-7B-Chat regarding the catastrophic forgetting problem on the general capabilities evaluation datasets in the main text, Table \ref{tab: main cf results}. For the results of other models, please refer to Appendix F. Compared to LLMs fine-tuned on the entire dataset, \textbf{GrADS brings substantial mitigation on CF}, i.e. 82.2\%, 79.5\%, 41.8\%, 104.8\%, 70.4\% improvements for C-Eval, GSM8K, ALPACA Instruct, Typical Safety, and Instruct Attack. 

From the domain perspective, we observe improvements of 79.3\%, 28.8\%, and 112.5\% on Medical, Legal, and Financial, respectively. Nevertheless, in medical (20k) and financial domains (40k) with larger training volumes, the gain of GrADS in alleviating CF problems is extremely significant.

\begin{figure*}[!ht]
    \centering
    \includegraphics[width=0.8\textwidth]{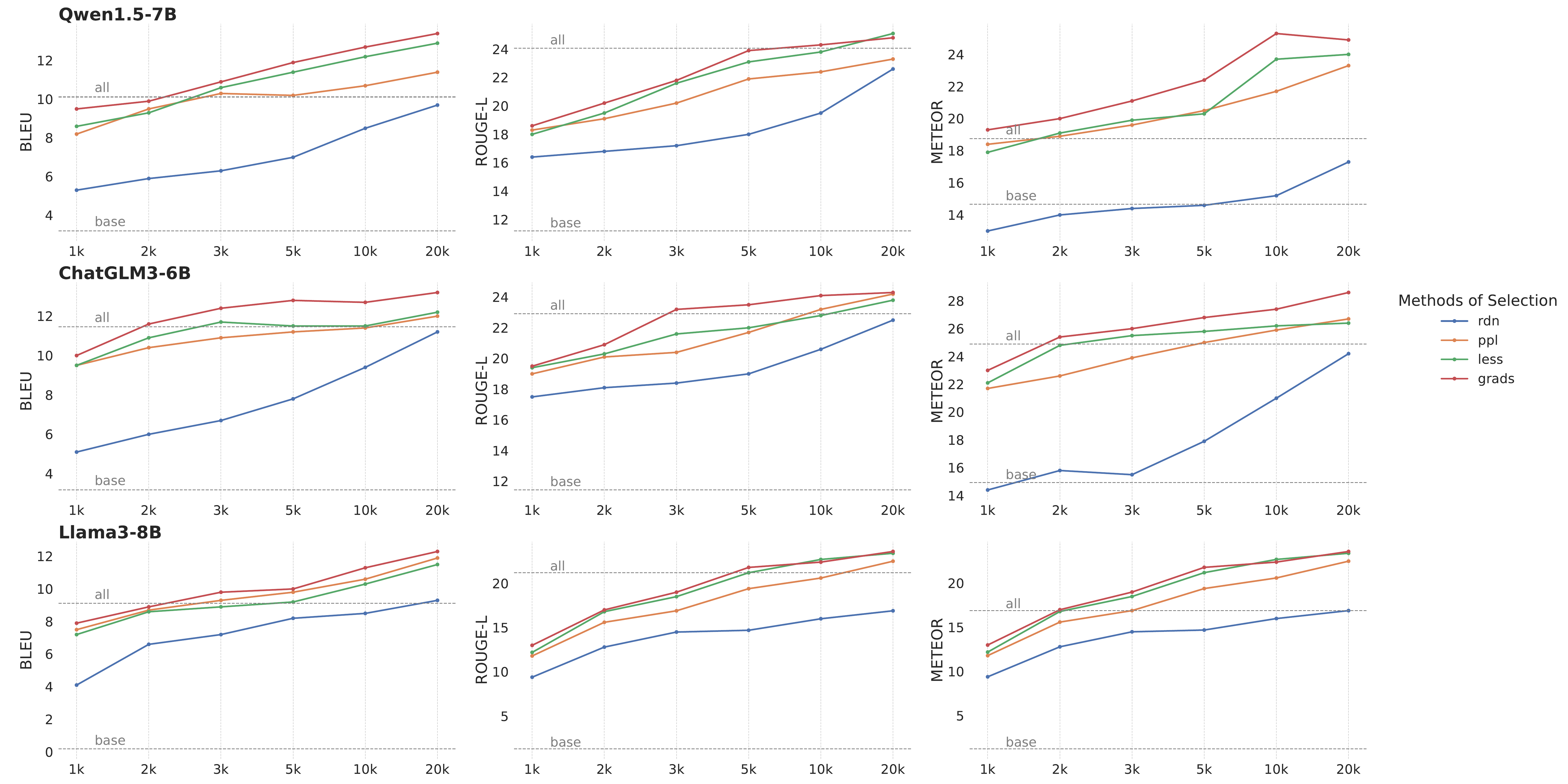}
    \caption{Experiments of fine-tuning Qwen1.5-7B, ChatGLM3-6B, Llama3-8B with subsets of different sizes selected from FinQA. Baselines 'base' and 'all' represent performances of the base models without SFT and the models fine-tuned on the entire data.}
    \label{fig:RQ2 result}
\end{figure*}

\subsection{Indepth Analysis}\label{subsec: indepth}
\subsubsection{RQ1: GrADS Generalizability}
The scaling law indicates that Larger models are significantly more sample-efficient \cite{kaplan2020scaling, zhang2024scaling}, so it is crucial to investigate whether GrADS is still valid in larger LLMs. Therefore, we selected Qwen1.5-14B and FinQA as our illustrative case. Meanwhile, to validate the transferability of GrADS, we initially train on Qwen1.5-1.8B for one epoch to acquire gradients for each instance and subsequently employ GrADS and other gradient-based baseline methods for data selection. Thereafter, the selected subdata is fine-tuned on Qwen1.5-14B. We present more experiments regarding the transferability of GrADS across various LLMs in Appendix C.

The results presented in Table \ref{tab: 14b results} demonstrate that GrADS not only \textbf{remains effective for larger LLMs} such as Qwen1.5-14B but also \textbf{can be applied across different LLMs}. Both findings highlight the strong generalizability of the GrADS method, offering exciting insights for researchers in the field of LLMs.

For instance, when confronted with voluminous training data, researchers can first leverage smaller LLMs, applying the GrADS strategy, before refining with relatively more efficacious larger LLMs. This strategy optimizes resource utilization, enabling the attainment of superior model performance while mitigating computational expenses – a pivotal consideration in large-scale machine-learning endeavors.

\subsubsection{RQ2: GrADS Robustness}
This section extends the main results by selecting 1k, 2k, 3k, 5k, and 10k training instances from FinQA with GrADS alongside other baseline approaches. These new experiments extend our prior analysis that was based on a 20k (50\%) selection, offering a broader perspective on GrADS' performance across varying data volumes. The experiment results are provided in Figure \ref{fig:RQ2 result}.

The results in Figure \ref{fig:RQ2 result} indicate that \textbf{the advantage of GrADS becomes even more evident in identifying subsets of smaller proportions}. In most cases, with merely 2.5\%-5\% (1k-2k) training instances, the GrADS has obtained comparable performance with those on full dataset. This finding holds immense implications for practical applications, showcasing a staggering efficiency-cost benefit ratio that could significantly transform the landscape of data utilization in language model tuning.

\section{Conclusion}
In this paper, to improve the fine-tuning efficiency and mitigate catastrophic forgetting simultaneously, we develop an adaptive gradient-aware data selection method, GrADS. Inspired by the insight that not all training data are helpful, GrADS integrates gradients extracted from the Embedding layer and LM Head layer and introduces self-guided criteria embracing statistic distributions to recognize the model's most desired data. Experimental results carried out on various LLMs and domain-specific datasets provide empirical evidence of the efficacy, efficiency, and cost-effectiveness of GrADS. Apart from extraordinary performance on domain-specific specialities, GrADS substantially mitigates catastrophic forgetting to preserve the general capabilities that the base LLMs mastered. Moreover, extensive analyses reveal that GrADS is also valid in the learning process of LoRA training, and can be scaled up to larger LLMs, delineating its great versatility and potential for generalizability.

\section*{Limitations}
In this paper, we introduce the GrADS method, which aims to enhance the efficiency of domain-specific fine-tuning. While extensive experiments validate the effectiveness of GrADS, our implementation was constrained by computational resource limitations, preventing us from applying GrADS to larger-scale language models (LLMs) with parameter sizes of 30B or 72B. Nevertheless, our focus primarily lies in resource-constrained scenarios; thus, experiments conducted with models ranging from 1.8B to 14B parameters are deemed sufficiently informative for our study. Investigations of GrADS on larger LLMs can be considered for future research endeavors.

\bibliography{custom}

\appendix

\begin{table*}[!ht]
    \centering
    \footnotesize
    \begin{tabular}{cl|m{0.8cm}<{\centering}m{0.8cm}<{\centering}m{1.1cm}<{\centering}|m{0.8cm}<{\centering}m{0.8cm}<{\centering}m{1.1cm}<{\centering}|m{0.8cm}<{\centering}m{0.8cm}<{\centering}m{1.1cm}<{\centering}}
    % \begin{tabular}{cl|ccc|ccc|ccc}
    \toprule
         \multirow{2}{*}{\textbf{Base Model}}& \multirow{2}{*}{\textbf{Method}} & \multicolumn{3}{c|}{\textbf{CMedQA}} & \multicolumn{3}{c|}{\textbf{LawQA}} & \multicolumn{3}{c}{\textbf{FinQA}} \\
         &  & BLEU & ROUGE & METEOR & BLEU & ROUGE &	METEOR & BLEU & ROUGE & METEOR \\
         \midrule
\multirow{7}{*}{\shortstack{Qwen1.5-1.8B}}&base&1.547& 8.228 & 11.169 & 9.860 & 19.178 & 24.505 & 1.888 & 7.911 & 12.061 \\
&all & 3.339 & 16.318 & 11.792 & 15.606 & \underline{26.124} & 25.950 & 9.358 & 21.014 & 17.127\\
\cmidrule(lr){2-11} 
&rdn & 3.515 & 16.005 & 11.233 & 14.973 & 25.587 & 24.653 & 8.491 & 20.726 & 16.440\\
&ppl & 4.147 & \underline{17.074} & 13.996 & 16.735 & 26.024 & 30.613 & 11.186 & \underline{21.673} & 20.360 \\
&less & \underline{4.392} & 16.950 & \underline{14.011} & \underline{17.067} & 25.962 & \underline{31.116} & 11.031 & 21.144 & \underline{20.468} \\
&grads & \textbf{4.852} & \textbf{18.218} & \textbf{14.439} & \textbf{18.754} & \textbf{26.339} & \textbf{33.688} & \textbf{11.875} & \textbf{22.106} & \textbf{21.732} \\
\midrule
\multirow{7}{*}{\shortstack{Qwen1.5-7B}}&base&2.627& 12.180 & 10.860 & 9.066 & 20.050 & 21.392 & 3.188 & 11.194 & 14.669 \\
&all & 3.813 & 17.327 & 12.276 & 16.090 & \underline{27.603} & 27.472 & 10.120 & 24.067 & 18.757\\
\cmidrule(lr){2-11} 
&rdn & 3.548 & 16.776 & 11.954 & 15.856 & 27.288 & 26.810 & 9.686 & 22.621 & 17.276\\
&ppl & \underline{4.832} & 18.215 & \textbf{15.012} & \underline{18.874} & 27.229 & \textbf{35.680} & \underline{11.302} & 22.276 & \underline{22.295} \\
&less & 4.711 & \textbf{18.976} & 13.690 & 17.822 & 27.144 & 29.539 & 10.157 & \underline{22.531} & 21.464 \\
&grads & \textbf{5.012} & \underline{18.664} & \underline{14.787} & \textbf{19.914} & \textbf{29.018} & \underline{34.457} & \textbf{12.782} & \textbf{24.064} & \textbf{23.980} \\
\midrule
\multirow{7}{*}{\shortstack{Qwen1.5-14B}}&base&2.934& 12.338 & 11.458 & 9.968 & 20.356 & 25.472 & 3.308 & 11.443 & 14.954 \\
&all & 4.034 & 17.577 & 12.760 & 16.688 & 27.763 & 28.381 & 12.518 & \underline{25.451} & 20.183\\
\cmidrule(lr){2-11} 
&rdn & 3.738 & 17.224 & 12.278 & 16.169 & 27.234 & 27.758 & 10.674 & 23.758 & 18.337\\
&ppl & 5.038 & 18.215 & \underline{15.339} & \underline{19.359} & 28.080 & \textbf{35.954} & 11.913 & 24.272 & 22.136 \\
&less & \underline{5.214} & \textbf{19.565} & 14.874 & 18.920 & 28.352 & 32.826 & \underline{13.549} & 24.877 & \underline{24.863} \\
&grads & \textbf{5.766} & \underline{19.018} & \textbf{15.862} & \textbf{20.314} & \textbf{30.523} & \underline{35.877} & \textbf{14.169} & \textbf{25.844} & \textbf{25.739} \\
\midrule
\multirow{7}{*}{\shortstack{ChatGLM3-6B}}&base&2.568 & 11.274 & 15.634 & 7.966 & 19.733 & 19.011 & 3.174 & 11.437 & 14.926 \\
&all & 4.297 & 17.432 & 16.722 & 16.673 & \textbf{28.016} & 28.519 & 11.454 & 22.918 & 24.898\\
\cmidrule(lr){2-11} 
&rdn & 4.512 & 16.674 & 16.482 & 16.453 & 27.576 & 27.864 & 11.216 & 22.450 & 24.233\\
&ppl & \underline{5.035} & 17.765 & \underline{17.930} & \underline{18.886} & 27.569 & \underline{31.783} & \underline{12.295} & 22.719 & \underline{26.327} \\
&less & 4.888 & \underline{18.026} & 17.651 & 17.379 & 27.468 & 29.136 & 11.843 & \underline{23.170} & 25.089 \\
&grads & \textbf{5.656} & \textbf{18.375} & \textbf{19.016} & \textbf{19.918} & \underline{27.779} & \textbf{34.154} & \textbf{13.328} & \textbf{24.434} & \textbf{27.977} \\
\midrule
\multirow{7}{*}{\shortstack{Llama3-8B}}&base& 0.026 & 0.249 & 0.291 & 0.259 & 1.905 & 2.164 & 0.178 & 1.293 & 1.225 \\
&all & 3.332 & 16.415 & 11.061 & 15.272 & 24.301 & 27.033 & 9.116 & 21.190 & 16.913\\
\cmidrule(lr){2-11} 
&rdn & 3.265 & 15.884 & 10.798 & 15.552 & 24.688 & 26.476 & 9.337 & 22.654 & 16.870 \\
&ppl & \underline{4.365} & 17.328 & \textbf{14.426} & \underline{18.225} & 25.964 & \underline{32.387} & \underline{11.454} & \underline{21.998} & 22.759 \\
&less & 4.186 & \underline{17.684} & 13.631 & 17.271 & \textbf{26.754} & 31.850 & 11.048 & 21.753 & \underline{22.833} \\
&grads & \textbf{4.774} & \textbf{18.125} & \underline{14.116} & \textbf{18.941} & \underline{26.376} & \textbf{33.385} & \textbf{12.028} & \textbf{23.366} & \textbf{23.300} \\
\bottomrule
    \end{tabular}
    \caption{Experiment results of implementing GrADS with \textbf{Qwen1.5-1.8B}, and leverage the selected data for SFT on \textbf{Qwen1.5-1.8B} itself and \textbf{other larger LLMs}. We select 50\% of data for training with \textit{rdn}, \textit{ppl}, \textit{less} and \textit{grads}.}
    \label{tab:transfer results}
\end{table*}

\section{Consistency between GPT-4o and Human Evaluation}
To ensure the validity of GPT-4o's assessments, we conduct supplementary manual evaluations. For the GPT-4o's response quality scores in the main results, we sampled 200 question-answer pairs rated by GPT-4o and enlisted the expertise of three professional data annotators to independently score the response quality on a scale of 1 to 5, in alignment with GPT-4o's scoring criteria. Subsequently, we calculated the average score for the three individuals and applied rounding to the nearest integer. We find that there are 147 samples for which the scores given by GPT-4o completely align with those of the annotators, and the scores from both sides yields a Pearson product-moment correlation coefficient of 0.79. This indicates a substantial agreement between the GPT-4o and human evaluations of response quality. 

Meanwhile, for the safety judgement of catastrophic forgetting experiments in main results, we also sample 200 instances for each category, i.e., Typical Safety and Instruction Attack, respectively. With the same approach, we measure the consistency between GPT-4o's judgments and the three annotators evaluations. We obtain a correlation coefficient score of 0.879 for Typical Safety and a correlation coefficient score of 0.815 for Instruction Attack.

\begin{table*}[!ht]
    \centering
    \footnotesize
    \begin{tabular}{cl|m{0.8cm}<{\centering}m{0.8cm}<{\centering}m{1.1cm}<{\centering}|m{0.8cm}<{\centering}m{0.8cm}<{\centering}m{1.1cm}<{\centering}|m{0.8cm}<{\centering}m{0.8cm}<{\centering}m{1.1cm}<{\centering}}
    % \begin{tabular}{cl|ccc|ccc|ccc}
    \toprule
         \multirow{2}{*}{\textbf{Method}}& \multirow{2}{*}{\textbf{Ablation}} & \multicolumn{3}{c|}{\textbf{CMedQA}} & \multicolumn{3}{c|}{\textbf{LawQA}} & \multicolumn{3}{c}{\textbf{FinQA}} \\
         &  & BLEU & ROUGE & METEOR & BLEU & ROUGE &	METEOR & BLEU & ROUGE & METEOR \\
         \midrule
\multirow{13}{*}{\shortstack{GrADS}}&base&2.627& 12.180 & 10.860 & 9.066 & 20.050 & 21.392 & 3.188 & 11.194 & 14.669 \\
&all & 3.813 & 17.327 & 12.276 & 16.090 & 27.603 & 27.472 & 10.120 & 24.067 & 18.757\\
&rdn & 3.548 & 16.776 & 11.954 & 15.856 & 27.288 & 26.810 & 9.686 & 22.621 & 17.276\\
\midrule			
&w/o lmhead & 4.435 & 17.019 & 14.455 & 18.866 & 25.759 & 33.887 & 11.945 & \textbf{25.012} & 23.130 \\
&w/o embed & \underline{5.011} & \underline{17.875} & 14.986 & 19.305 & 26.874 & 32.491 & \underline{12.455} & 24.170 & \underline{23.843} \\
&top grad & 2.986 & 15.874 & 10.039 & 14.012 & 25.183 & 23.456 & 6.758 & 20.417 & 14.009 \\
&tail grad & 4.736 & 16.689 & \underline{15.006} & \underline{19.424} & 26.780 & \underline{33.699} & 12.274 & 23.510 & 23.356 \\
&mid grad & 4.630 & 17.492 & 14.890 & 17.764 & \underline{27.733} & 33.383 & 10.429 & 22.304 & 22.285 \\
& weight & 4.832 & 17.316 & 14.284 & 19.259 & 26.671 & 33.034 & 12.218 & 23.313 & 23.409\\
& weightr & 4.727 & 17.134 & 13.855 & 18.736 & 26.033 & 32.682 & 11.769 & 22.237 & 23.030 \\
\midrule
\multirow{1}{*}{\shortstack{GRADS}}
& ours & \textbf{5.372} & \textbf{18.496} & \textbf{15.396} & \textbf{20.270} & \textbf{28.026} & \textbf{35.985} & \textbf{13.364} & \underline{24.822} & \textbf{24.872} \\
\bottomrule
    \end{tabular}
    \caption{Ablation Study. We select 50\% of data for training except for \textit{base} and \textit{all}.}
    \label{tab:ablation study}
\end{table*}

\begin{figure*}[!ht]
    \centering
    \includegraphics[width=1.0\textwidth]{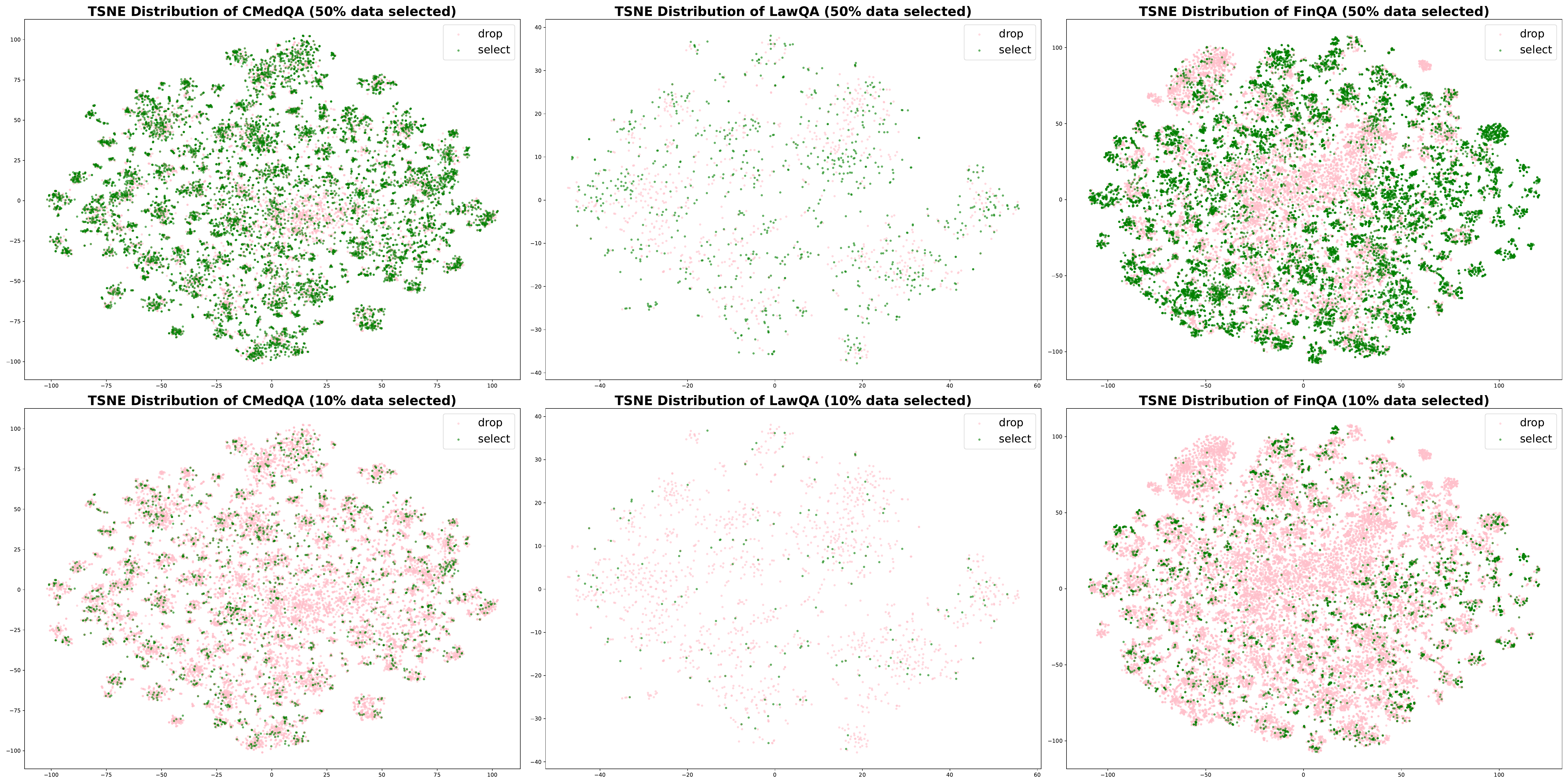}
    \caption{Semantic distribution of training instances. The green dots indicate selected instances whereas the red dots indicate dropped instance.}
    \label{fig:diversity study}
\end{figure*}

% \begin{figure}[t]
%     \centering
%    \includegraphics[width=\linewidth]{figure/Appendix_D_plot.pdf}
%     \caption{Semantic distribution of training instances. The green dots indicate selected instances whereas the red dots indicate dropped instance.}
%     \label{fig:diversity study}
% \end{figure}

\section{GrADS Transferability on Different LLMs}

In this section, we adhere to the setting in RQ1 which initially train on Qwen1.5-1.8B for one epoch to acquire gradients and employ Qwen1.5-1.8B itself or other LLMs for fine-tuning. Our findings in Table \ref{tab:transfer results} reveal that data selected using Qwen1.5-1.8B in conjunction with GrADS not only prove effective for larger LLMs of the same architecture (Qwen-1.5-7B and Qwen1.5-14B), but also yield substantial improvements for larger LLMs of different architectures, including ChatGLM3-6B and Llama3-8B. This experimentation \textbf{further validates the transferability of the GrADS methodology}.

\section{Ablation Study}

In this section, we conducted an ablation study using Qwen1.5-7B as a case example across three domains. Specifically, we examined the impacts of incorporating only the gradients from the Embed Layer or only from the LM Head Layer, namely \textit{w/o lmhead} and \textit{w/o embed}. Additionally, we also investigate how LLMs perform when they are trained on data selected from the half with the smallest gradients (\textit{tail}), the largest half (\textit{top}), and the middle half (\textit{mid}). Meanwhile, as in GrADS we add the gradients from the LM Head Layer and Embed Layer directly, and we also explore substitute integration methods. In Table\ref{tab:ablation study}, \textit{weight} refers to the gradients from the Embed Layer and LM Head Layer that are normalized and summed to derive a gradient distribution. Besides, \textit{weightr} entails ranking the gradients of each instance from the Embed Layer and LM Head Layer in descending order and summing their ranks' reciprocals to obtain the distribution. Subsequently, both \textit{weight} and \textit{weightr} utilize the same data selection criteria as GrADS. 

The experimental results indicate that our original GrADS consistently achieves optimal or sub-optimal performance, thereby validating the rationale behind our methodological design.

\begin{table*}[!ht]
    \centering
    \footnotesize
    \begin{tabular}{cl|cc|ccc|cc|c|c}
    \toprule
         \multirow{2}{*}{\textbf{Domain}}& \multirow{2}{*}{\textbf{Method}} & \multicolumn{2}{c|}{\textbf{C-Eval}} & \multicolumn{3}{c|}{\textbf{GSM8k}} & \multicolumn{2}{c|}{\textbf{ALPACA}} & \multicolumn{1}{c|}{\textbf{Safety}} & \multicolumn{1}{c}{\textbf{Attack}} \\
         &  & Acc & Instruct & Acc &	BLEU & ROUGE & BLEU & ROUGE & Acc & Acc \\
         \midrule
\multirow{6}{*}{\shortstack{CMedQA}}&base&54.360& 73.328 & 46.020 & 18.593 & 33.002 & 16.362 & 28.321 & 44.681 & 50.686 \\
\midrule
&all & 8.794 & 5.523 & 3.942 & 4.271 & 16.108 & 4.782 & 16.971 & 16.783 & 23.068 \\
% \cmidrule(lr){2-11} 
&rdn & 15.480 & 18.023 & 6.823 & 5.811 & 18.120 & 6.636 & 19.507 & 16.879 & 27.863 \\
% &top50 & 14.389 & 17.297 & 5.080 & 4.311 & 16.087 & 5.197 & 17.812 & 14.132 & 19.164 \\
% &tail50 & \underline{16.119} & \underline{18.305} & \underline{9.704} & \textbf{7.025} & \textbf{20.333} & \textbf{8.231} & \textbf{20.987} & \textbf{27.834} & \textbf{35.233} \\
&grads & \textbf{16.480} & \textbf{19.695} & \textbf{10.008} & \textbf{6.468} & \textbf{19.098} & \textbf{7.810} & \textbf{20.765} & \textbf{25.679} & \textbf{31.484} \\
\midrule
\multirow{3}{*}{\shortstack{LawQA}}
&all & 30.451 & 25.363 & 34.572 & 14.833 & 28.227 & 13.966 & 25.468 & 26.137 & 40.017 \\
% \cmidrule(lr){2-11} 
&rdn &32.756 & \textbf{38.227} & 36.012 & \textbf{16.103} & \textbf{29.974} & \textbf{15.284} & \textbf{27.242} & 28.995 & 41.983\\
% &top50 & 30.977 & 29.288 & 34.723 & 14.245 & 27.353 & 13.461 & 24.416 & 23.341 & 37.615 \\
% &tail50 & 32.669 & 37.631 & \underline{36.922} & 15.561 & \underline{29.144} & \textbf{14.825} & 26.173 & \underline{35.779} & \underline{40.885} \\
&grads & \textbf{33.717} & 37.974 & \textbf{37.225} & 15.709 & 29.080 & 13.637 & 26.671 & \textbf{37.681} & \textbf{42.036} \\
\midrule
\multirow{3}{*}{\shortstack{FinQA}}&all & 13.953 & 2.947 & 1.365 & 1.418 & 9.869 & 2.832 & 12.448 & 7.416 & 11.678\\
% \cmidrule(lr){2-11} 
&rdn & 18.823 & 5.794 & 2.578 & 2.032 & 11.751 & 3.638 & 14.495 & 10.861 & 14.884 \\
% &top50 & 16.497 & 3.329 & 1.971 & 0.946 & 9.499 & 2.938 & 13.073 & 8.366 & 10.550 \\
% &tail50 & 18.753 & \underline{7.742} & \textbf{4.473} & \textbf{4.228} & \textbf{16.460} & \textbf{4.999} & \underline{15.773} & \underline{19.451} & \textbf{25.276} \\
&grads & \textbf{26.017} & \textbf{19.089} & \textbf{4.250} & \textbf{3.717} & \textbf{15.649} & \textbf{4.911} & \textbf{16.776} & \textbf{21.025} & \textbf{24.481} \\
\bottomrule
    \end{tabular}
    \caption{Supplementary experiments of \textbf{Catastrophic Forgetting} on \textbf{ChatGLM3-6B}. We select 50\% of data for training with \textit{rdn} and \textit{grads}}
    \label{tab:glm6b cf results}
\end{table*}

\begin{table*}[!ht]
    \centering
    \footnotesize
    \begin{tabular}{cl|cc|ccc|cc|c|c}
    \toprule
         \multirow{2}{*}{\textbf{Domain}}& \multirow{2}{*}{\textbf{Method}} & \multicolumn{2}{c|}{\textbf{C-Eval}} & \multicolumn{3}{c|}{\textbf{GSM8k}} & \multicolumn{2}{c|}{\textbf{ALPACA}} & \multicolumn{1}{c|}{\textbf{Safety}} & \multicolumn{1}{c}{\textbf{Attack}} \\
         &  & Acc & Instruct & Acc &	BLEU & ROUGE & BLEU & ROUGE & Acc & Acc \\
         \midrule
\multirow{6}{*}
{\shortstack{CMedQA}}&base&46.657& 97.832 & 58.226 & 3.903 & 16.089 & 3.229 & 7.443 & 27.150 & 44.167 \\
\midrule
&all & 0.291 & 0.000 & 0.682 & 0.894 & 9.215 & 0.861 & 9.317 & 7.061 & 6.333 \\
&rdn & 0.390 & 0.036 & 0.076 & 0.777 & 8.749 & 0.860 & 9.310 & 10.535 & 5.583 \\
&grads & \textbf{0.509} & \textbf{0.073} & \textbf{0.758} & \textbf{1.679} & \textbf{10.827} & \textbf{1.131} & \textbf{10.800} & \textbf{14.672} & \textbf{10.250} \\
\midrule
\multirow{3}{*}{\shortstack{LawQA}}
&all & 3.634 & 3.343 & 2.729 & 4.438 & 13.415 & 4.624 & 14.708 & 15.643 & 32.333 \\
&rdn & 4.506 & 0.727 & 3.033 & 4.105 & 12.859 & 4.298 & 13.869 & 22.714 & 37.250 \\
&grads & \textbf{4.869} & \textbf{6.017} & \textbf{4.701} & \textbf{6.311} & \textbf{15.585} & \textbf{5.326} & \textbf{16.081} & \textbf{25.143} & \textbf{41.583} \\
\midrule
\multirow{3}{*}{\shortstack{FinQA}}&all & 0.363 & 1.817 & 0.227 & 0.491 & 6.985 & 0.558 & 10.297 & 1.500 & 2.750 \\
&rdn & 1.438 & 9.084 & 0.455 & 0.683 & 7.019 & 0.667 & 8.091 & \textbf{2.571} & \textbf{3.917} \\
&grads & \textbf{6.541} & \textbf{21.148} & \textbf{0.607} & \textbf{1.019} & \textbf{7.514} & \textbf{0.739} & \textbf{8.473} & 1.929 & 3.833 \\
\bottomrule
    \end{tabular}
    \caption{Supplementary experiments of \textbf{Catastrophic Forgetting} on \textbf{Llama3-8B}. We select 50\% of data for training with \textit{rdn} and \textit{grads}}
    \label{tab:llama8b cf results}
\end{table*}

\section{Data Diversity}
One concern is that selecting data based on the highest probability density might compromise the diversity of the chosen dataset, an aspect that is essential for effective large language model (LLM) training. Therefore, in this section, we apply GrADS with Qwen1.5-7B for data selection across three domains. To obtain the semantic distribution of training instances, we apply Text\_Embedding\_V3 \footnote{https://www.alibabacloud.com/help/en/model-studio/developer-reference/text-embedding-synchronous-api} for embedding representation and TSNE \cite{van2008visualizing} technique for dimensionality reduction and visualization. 

The results illustrated in Figure \ref{fig:diversity study} suggest that the probability density of gradients has few relevance to semantic meanings. Notably, the data selected by GrADS maintain considerable diversity, regardless of the situation of 50\% or 10\% selection. As we have discussed in the Introduction section, LLMs can perform like absolutely rational college students who select courses they need not just what they like.

\section{Supplementary Experiments of Catastrophic Forgetting}

In this section, we provide supplementary experimental results regarding catastrophic forgetting problem. Table \ref{tab:glm6b cf results} and table \ref{tab:llama8b cf results} illustrate the results of ChatGLM3-6B and Llama3-8B, which validate that GrADS not only substantially alleviate catastrophic forgetting for Qwen1.5-7B, but also for ChatGLM3-6B and Llama3-8B.

\section{Supplementary Experiments of LoRA Tuning}

Apart from full parameter fine-tuning, we also investigate how GrADS would facilitate LoRA tuning. Table \ref{tab:lora main results} provides the results of LoRA tuning whereas table \ref{tab: lora qwen7b cf results}, table \ref{tab: lora chatglm6b cf results}, and table \ref{tab: lora llama8b cf results} provide the results of the catastrophic forgetting problem of Qwen1.5-7B, ChatGLM3-6B, and Llama3-8B after LoRA tuning, respectively.

Those experiments validate GrADS's effectiveness across full parameter fine-tuning and LoRA tuning. In the meantime, for those who seeking a balance between domain capabilities and general capabilities (less catastrophic forgetting), the combination of GrADS and LoRA tuning should be a good choice.

\begin{table*}[!ht]
    \centering
    \footnotesize
    \begin{tabular}{cl|m{0.8cm}<{\centering}m{0.8cm}<{\centering}m{1.1cm}<{\centering}|m{0.8cm}<{\centering}m{0.8cm}<{\centering}m{1.1cm}<{\centering}|m{0.8cm}<{\centering}m{0.8cm}<{\centering}m{1.1cm}<{\centering}}
    % \begin{tabular}{cl|ccc|ccc|ccc}
    \toprule
         \multirow{2}{*}{\textbf{Base Model}}& \multirow{2}{*}{\textbf{Method}} & \multicolumn{3}{c|}{\textbf{CMedQA}} & \multicolumn{3}{c|}{\textbf{LawQA}} & \multicolumn{3}{c}{\textbf{FinQA}} \\
         &  & BLEU & ROUGE & METEOR & BLEU & ROUGE &	METEOR & BLEU & ROUGE & METEOR \\
         \midrule
\multirow{9}{*}{\shortstack{Qwen1.5-7B}}&base & 2.627 & 12.180 & 10.860 & 9.066 & 20.050 & 21.392 & 3.188 & 11.194 & 14.669 \\
&all & 4.075 & \underline{17.739} & 12.966 & 14.580 & \textbf{27.382} & \underline{31.207} & 7.316 & 20.192 & 15.365 \\
\cmidrule(lr){2-11} 
&rdn & 3.839 & 17.219 & 12.250 & 14.293 & 25.581 & 27.769 & 6.372 & 19.415 & 14.108 \\
&bm25 & 3.555 & 16.875 & 11.208 & 12.937 & 24.837 & 25.981 & 5.709 & 18.793 & 12.283 \\
&dsir & 3.840 & 16.698 & 11.475 & 13.057 & 24.880 & 24.512 & 5.716 & 17.397 & 12.388 \\
&rds & 3.818 & 17.022 & 11.549 & 13.235 & 24.320 & 23.898 & 6.875 & 20.051 & 12.648 \\
&ppl & 4.526 & 17.481 & 13.569 & 14.862 & 24.383 & 25.485 & 7.769 & \underline{20.651} & 16.866 \\
&less & \underline{4.757} & 17.596 & \textbf{14.984} & \underline{16.012} & 26.057 & 30.136 & \underline{7.892} & 20.135 & \underline{17.200} \\
&grads & \textbf{5.018} & \textbf{18.243} & \underline{14.696} & \textbf{17.963} & \underline{26.755} & \textbf{32.802} & \textbf{9.103} & \textbf{21.154} & \textbf{18.848} \\
\midrule
\multirow{9}{*}{\shortstack{ChatGLM3-6B}}&base & 2.568 & 11.274 & 10.634 & 7.966 & 19.733 & 19.011 & 3.174 & 11.437 & 14.926 \\
&all & 3.551 & 15.960 & 12.124 & \textbf{12.903} & \textbf{22.514} & 23.174 & 8.047 & \underline{20.820} & 17.159 \\
\cmidrule(lr){2-11} 
&rdn & 3.498 & 15.824 & 11.970 & 10.010 & 20.038 & 21.166 & 8.155 & 19.896 & 17.032 \\
&bm25 & 3.539 & 16.296 & 12.035 & 10.457 & 20.745 & 20.899 & 8.100 & 20.043 & 16.747 \\
&dsir & 3.667 & 16.187 & 11.892 & 9.964 & 20.819 & 20.451 & 8.269 & 19.803 & 16.760 \\
&rds & 3.256 & 15.517 & 11.389 & 9.854 & 19.899 & 20.016 & 7.079 & 19.266 & 16.148 \\
&ppl & \underline{4.286} & \underline{17.536} & 13.492 & 11.914 & 20.188 & 22.358 & 8.177 & 20.375 & 16.658 \\
&less & 3.932 & 16.774 & \underline{13.758} & 11.616 & 21.089 & 21.648 & \underline{8.524} & 20.793 & \underline{17.617} \\
&grads50 & \textbf{4.483} & \textbf{18.216} & \textbf{14.447} & \underline{12.724} & \underline{22.214} & \textbf{23.857} & \textbf{8.896} & \textbf{21.301} & \textbf{17.966} \\
\midrule
\multirow{9}{*}{\shortstack{Llama3-8B}}&base & 0.026 & 0.249 & 0.291 & 0.259 & 1.905 & 2.164 & 0.178 & 1.293 & 1.225 \\
&all & 3.138 & \textbf{16.695} & 11.782 & \underline{16.125} & \textbf{25.588} &   28.327 &  \underline{9.336} &  \underline{22.480} &  18.751\\
\cmidrule(lr){2-11} 
&rdn & 2.851 & 16.030 & 10.956 &   14.478 &   24.515&   27.160 &   8.931&   21.267 &  16.922\\
&bm25 & 2.543 & 15.381 & 9.075 &   13.308&   21.629&   25.584 &   7.856&   20.639 &  14.487 \\
&dsir & 2.738 & 15.683 & 10.719 &   13.985 &   24.205 &   26.650 &   8.857 &   20.977 &  17.356 \\
&rds & 2.918 & 15.984 & 10.270 &   14.041&   23.388&   26.986 &   8.844 &   20.074 &  17.706 \\
&ppl & 3.326 & \underline{16.540} & 12.016 &   15.427&   23.958 &   \underline{29.836} &   9.328&   20.890 &  \textbf{19.027} \\
&less & \textbf{3.517} & 16.310 & \underline{12.022} &   15.811 &   24.018 &   28.895 &   9.085 &   20.228 &  18.750 \\
&grads & \underline{3.446} & 16.019 & \textbf{12.527} &  \textbf{16.475} &  \underline{25.487} &   \textbf{30.699} &  \textbf{9.919} &  \textbf{22.807} &  \underline{18.940} \\
\bottomrule
    \end{tabular}
    \caption{Supplementary experiments of \textbf{LoRA} tuning. \textit{base} denotes no further training implemented, \textit{all} denotes full dataset, and otherwise we select 50\% of the data for training.}
    \label{tab:lora main results}
\end{table*}

\section{Baseline Illustration}
We present a brief introduction of our baselines in this section. \textbf{BM25} \cite{robertson2009probabilistic} featurizes examples by their word frequency statistics (i.e., TF-IDF) to rank the training instances, and select the top k\% of the training instances with the highest scores to construct Dtrain. \textbf{DSIR} \cite{xie2023data} uses n-gram features to weight candidate training data D. We resample k\% of the training instances according to the importance weights. \textbf{RDS} (Representation-based Data Selection) \cite{zhang2018unreasonable, hanawa2020evaluation} uses the model’s hidden representations as features for data selection. We follow the settings in \citet{xia2024less}, which computes the similarity score using Equation (2) of \citet{xia2024less} but replace the gradient features with the final layer representations of the last token of each sequence. \textbf{LESS} (Low-rank gradiEnt Similarity Search) \cite{xia2024less} utilizes gradients as well and selects training instances based on their similarity to few-shot examples embodying a specific capability.

\section{Implementation Details}
Our experiment is conducted on 8 A100 GPUs, each with 80G memories. All experiments are conducted with LLaMA-Factory\footnote{https://github.com/hiyouga/LLaMA-Factory/tree/main} training architecture and deepspeed\_z3. For all methods, we set the learning rate of 3e-5, warmup ratio of 0.1, and batch size of 8. Regarding LLMs’ API, we adopt GPT-4o. For LoRA experiments, the rank is set to 16. For all randomly selected data, we set the random seed of 42. To maintain some basic instruction following capabilities for more precise evaluation (especially for \textit{rdn} and \textit{all}), for all catastrophic forgetting related experiments, we only report the score on the test set after 1 training epoch. For the rest of the experiments, we report the average scores on the test set after the training epochs of 1, 2, and 3.

\begin{table*}[!ht]
    \centering
    \footnotesize
    \begin{tabular}{cl|cc|ccc|cc|c|c}
    \toprule
         \multirow{2}{*}{\textbf{Domain}}& \multirow{2}{*}{\textbf{Method}} & \multicolumn{2}{c|}{\textbf{C-Eval}} & \multicolumn{3}{c|}{\textbf{GSM8k}} & \multicolumn{2}{c|}{\textbf{ALPACA}} & \multicolumn{1}{c|}{\textbf{Safety}} & \multicolumn{1}{c}{\textbf{Attack}} \\
         &  & Acc & Instruct & Acc &	BLEU & ROUGE & BLEU & ROUGE & Acc & Acc \\
         \midrule
\multirow{6}{*}{\shortstack{CMedQA}}&base& 65.189 & 87.427 & 55.497  &  14.967 & 29.207  & 15.097  & 27.529  & 43.807  & 51.365  \\
\midrule
&all & \textbf{35.512} & \textbf{42.124} & 22.592 & 7.368 & 23.634 & 5.681 & 19.265 &  23.087 & 31.415 \\
% \cmidrule(lr){2-11} 
&rdn & 29.420 & 29.940 & 33.131 & 10.441 & 27.717 & 7.044 & 21.033 & 28.596  & 37.847 \\
% &top50 & 37.741 & 47.399 & 31.311 & 8.881 & 26.047 & 5.371 & 18.705 &   &   \\
% &tail50 & 27.488 & 27.786 & 45.337 & 15.688 & 32.247 & 9.818 & 23.387 &   &   \\
&grads & 34.101 & 40.638 & \textbf{44.806} & \textbf{14.600} & \text{31.424} & \textbf{8.644} & \textbf{22.966} & \textbf{31.138}  & \textbf{42.636} \\
\midrule
\multirow{3}{*}{\shortstack{LawQA}}
&all& 34.323 & 32.615 & 53.373 & 14.408 & 28.252 & 14.031 & 26.393 &   28.650 &  41.684 \\
% \cmidrule(lr){2-11} 
&rdn & 39.673 & \textbf{39.598} & 53.146 & \textbf{14.841} & \textbf{29.121} & \textbf{14.970} & \textbf{27.211} & 33.757  & \textbf{50.220}  \\
% &top50 & 39.524 & 39.524 & 53.297 & 16.680 & 31.275 & 16.143 & 28.583 &   &   \\
% &tail50 & 41.976 & 42.199 & 54.510 & 13.227 & 26.900 & 13.843 & 25.412 &   &   \\
&grads & \textbf{41.307} & 38.484 & \textbf{53.980} & 12.970 & 26.678 & 13.504 & 25.160 & \textbf{35.766}  & 49.814  \\
\midrule
\multirow{3}{*}{\shortstack{FinQA}}&all & 48.365 & 68.870 & 17.664 & 3.528 & 18.015 & 4.178 & 15.660 &  17.174 & 24.269  \\
% \cmidrule(lr){2-11} 
&rdn & \textbf{50.817} & \textbf{70.653} & 20.849 & 3.976 & 19.035 & 4.356 & 16.408 &  21.235 &  33.471 \\
% &top50 & 52.674 & 73.476 & 18.650 & 3.373 & 18.160 & 4.007 & 15.486 &   &   \\
% &tail50 & 28.454 & 27.860 & 33.131 & 10.435 & 28.006 & 6.115 & 18.611 &   &   \\
&grads & 27.637 & 22.956 & \textbf{28.506} & \textbf{8.582} & \textbf{26.069} & \textbf{6.000} & \textbf{18.344} &  \textbf{27.451} &  \textbf{36.045} \\
\bottomrule
    \end{tabular}
    \caption{Supplementary experiments of \textbf{Catastrophic Forgetting} after \textbf{LoRA} tuning on \textbf{Qwen1.5-7B}. We select 50\% of data for training with \textit{rdn} and \textit{grads}}
    \label{tab: lora qwen7b cf results}
\end{table*}

\begin{table*}[!ht]
    \centering
    \footnotesize
    \begin{tabular}{cl|cc|ccc|cc|c|c}
    \toprule
         \multirow{2}{*}{\textbf{Domain}}& \multirow{2}{*}{\textbf{Method}} & \multicolumn{2}{c|}{\textbf{C-Eval}} & \multicolumn{3}{c|}{\textbf{GSM8k}} & \multicolumn{2}{c|}{\textbf{ALPACA}} & \multicolumn{1}{c|}{\textbf{Safety}} & \multicolumn{1}{c}{\textbf{Attack}} \\
         &  & Acc & Instruct & Acc &	BLEU & ROUGE & BLEU & ROUGE & Acc & Acc \\
         \midrule
\multirow{6}{*}{\shortstack{CMedQA}}&base&   54.360&   73.328&   46.020&   18.593&   33.002&   16.362&   28.321& 44.681 &  50.686 \\
\midrule
&all &   25.186&   28.232&   25.929&   11.616&   27.839&   11.079&   24.465& 25.318  &  33.572 \\
% \cmidrule(lr){2-11} 
&rdn &   \textbf{30.163} &   29.822&   30.857&   13.497&   29.993&   12.162&   25.685&  31.664 &  38.055 \\
% &top50 &   31.724&   41.530&   31.16&   13.166&   29.492&   11.810&   25.320&   &   \\
% &tail50 &   31.129&   42.348&   33.378&   14.865&   31.204&   12.931&   26.167&   &   \\
&grads &   28.678&   \textbf{40.416} &  \textbf{33.207} &   \textbf{14.455} &  \textbf{31.127} &  \textbf{12.426} &  \textbf{25.974} & \textbf{32.042}  &  \textbf{39.776} \\
\midrule
\multirow{3}{*}{\shortstack{LawQA}}
&all&   39.673&   60.327&   42.077&   17.661&   32.429&   14.814&   27.301&  30.285 &  40.069 \\
% \cmidrule(lr){2-11} 
&rdn &   39.004&   \textbf{64.859} &   43.821&   \textbf{17.929} &   \textbf{32.610} &   \textbf{15.249} &   \textbf{27.640} & 35.460  &  46.734 \\
% &top50 &   41.085&   66.790&   42.305&   18.050&   32.604&   15.216&   28.010&   &   \\
% &tail50 &   41.159&   67.236&   42.608&   17.913&   32.652&   14.773&   27.327&   &   \\
&grads &   \textbf{40.119} &   64.636&   \textbf{44.655} &   17.895&   32.517&   15.065&   27.446&  \textbf{37.261} &  \textbf{48.588} \\
\midrule
\multirow{3}{*}{\shortstack{FinQA}}&all &   23.031&   42.422&   22.214&   10.278&   26.400&   10.022&   22.785 &  16.292 &  25.106 \\
% \cmidrule(lr){2-11} 
&rdn &   28.158&   40.416&   \textbf{26.005} &   11.090&   27.576&   10.744&   23.197&  22.234 &  30.217 \\
% &top50 &   29.643&   52.229&   18.575&   7.158&   22.996&   8.536&   21.609&   &   \\
% &tail50 &   29.346&   46.731&   30.250&   13.147&   29.368&   12.055&   24.878&   &   \\
&grads &   \textbf{31.055} &   \textbf{45.840} &   24.867&   \textbf{11.863} &   \textbf{28.318} &   \textbf{12.641} &   \textbf{25.519} &  \textbf{28.656} &  \textbf{35.785} \\
\bottomrule
    \end{tabular}
    \caption{Supplementary experiments of \textbf{Catastrophic Forgetting} after \textbf{LoRA} tuning on \textbf{ChatGLM3-6B}. We select 50\% of data for training with \textit{rdn} and \textit{grads}}
    \label{tab: lora chatglm6b cf results}
\end{table*}

\begin{table*}[!ht]
    \centering
    \footnotesize
    \begin{tabular}{cl|cc|ccc|cc|c|c}
    \toprule
         \multirow{2}{*}{\textbf{Domain}}& \multirow{2}{*}{\textbf{Method}} & \multicolumn{2}{c|}{\textbf{C-Eval}} & \multicolumn{3}{c|}{\textbf{GSM8k}} & \multicolumn{2}{c|}{\textbf{ALPACA}} & \multicolumn{1}{c|}{\textbf{Safety}} & \multicolumn{1}{c}{\textbf{Attack}} \\
         &  & Acc & Instruct & Acc &	BLEU & ROUGE & BLEU & ROUGE & Acc & Acc \\
         \midrule
\multirow{6}{*}{\shortstack{CMedQA}}&base&   46.657&   97.832&   58.226&   3.903&   16.089&   3.299&   7.443 & 27.150  & 44.167  \\
\midrule
&all & 2.674 & 1.783 & 15.693 & 6.706 & 21.267 & 3.343 & 14.913 & 15.714  &  20.333 \\
% \cmidrule(lr){2-11} 
&rdn & 9.212 & 14.413 & 17.664 & 6.666 & 21.489 & 3.149 & 15.131 &  21.071 &  26.750 \\
% &top50 & 15.378 & 34.026 & 17.058 & 5.519 & 21.366 & 2.683 & 14.017 &   &   \\
% &tail50 & 5.720 & 8.098 & 29.492 & 10.014 & 26.372 & 5.806 & 19.154 &   &   \\
&grads & \textbf{9.509} & \textbf{18.127} & \textbf{23.730} & \textbf{8.970} & \textbf{25.342} & \textbf{5.402} & \textbf{18.493} & \textbf{22.286}  &  \textbf{29.167} \\
\midrule
\multirow{3}{*}{\shortstack{LawQA}}
&all&   39.376&   81.278&   54.814&   17.542&   \textbf{33.502} &   \textbf{10.398} &   \textbf{21.001} &  24.643 &  31.250\\
% \cmidrule(lr){2-11} 
&rdn &   \textbf{44.428} &   \textbf{92.422} &   55.800&   16.637&   32.829&   9.611&   19.168&  26.143 & 34.333  \\
% &top50 &   42.496&   93.611&   54.056&   16.439&   33.342&   11.033&   21.827&   &   \\
% &tail50 &   43.388&   91.456&   58.074&   17.421&   33.379&   8.921&   17.570&   &   \\
&grads &   43.908&   91.976&   \textbf{58.302} &   \textbf{17.660} &   33.339&   8.545&   17.339& \textbf{28.643}  &  \textbf{40.667} \\
\midrule
\multirow{3}{*}{\shortstack{FinQA}}&all &   22.140&   28.826&   30.857&   6.806&   25.282&   5.200&   17.511&  7.214 &  14.250 \\
% \cmidrule(lr){2-11} 
&rdn &   \textbf{28.158} &   \textbf{54.309} &   32.980&   7.548&   26.324&   5.536&   18.784& 11.857  &  19.417 \\
% &top50 &   22.288&   40.416&   26.384&   5.526&   22.858&   3.754&   15.389&   &   \\
% &tail50 &   18.276&   26.895&   32.980&   8.832&   27.390&   6.798&   19.479&   &   \\
&grads &   23.626&   35.364&   \textbf{34.117} &   \textbf{9.022} &   \textbf{27.687} &   \textbf{6.566} &   \textbf{19.203} &  \textbf{14.143} & \textbf{24.083}  \\
\bottomrule
    \end{tabular}
    \caption{Supplementary experiments of \textbf{Catastrophic Forgetting} after \textbf{LoRA} tuning on \textbf{Llama3-8B}. We select 50\% of data for training with \textit{rdn} and \textit{grads}}
    \label{tab: lora llama8b cf results}
\end{table*}

\end{document}